\documentclass[journal]{IEEEtran}
\usepackage{xcolor}
\usepackage{booktabs}
\usepackage{epsfig}
\usepackage{graphicx}
\usepackage{amsmath}

\usepackage{amssymb}

\usepackage{verbatim}

\usepackage{url}
\usepackage{tabularx}
\usepackage{multirow}
\usepackage{subfigure}
\usepackage{amsmath,dsfont}
\usepackage{array}
\usepackage{enumitem}

\usepackage{pifont}
\usepackage{threeparttable}
\usepackage{bbding}
\usepackage[utf8]{inputenc}
\usepackage{verbatim}

\usepackage{colortbl}
\definecolor{mygray}{gray}{.75}

\usepackage{xcolor}

\usepackage{bm}
\newcommand{\et}[2]{${#1}^{\pm{#2}}$}
\newcommand{\etb}[2]{$\mathbf{{#1}}^{\pm{#2}}$}
\newcommand{\ets}[2]{$\textcolor{blue}{{#1}}^{\pm{#2}}$}
\usepackage{multirow}

\usepackage[normalem]{ulem}

\usepackage{makecell}
\usepackage{multirow}
\usepackage[linesnumbered,ruled,vlined]{algorithm2e}
\usepackage{algorithmic}

\begin{document}
%

\title{Embracing Aleatoric Uncertainty: Generating Diverse 3D Human Motion}
%
%
%
 \author{Zheng~Qin,
        Yabing~Wang,
        Minghui Yang,
        Sanping Zhou,~\IEEEmembership{Member,~IEEE}
        Ming Yang,~\IEEEmembership{Member,~IEEE}
        Le~Wang,~\IEEEmembership{Senior Member,~IEEE}

	\thanks{

This work was supported in part by the National Key R\&D Program of China under Grant 2021YFB1714700, in part by NSFC under Grants 62088102, 62106192, and 12326608, in part by the Natural Science Foundation of Shaanxi Province under Grant 2022JC-41, and in part by Fundamental Research Funds for the Central Universities under Grant XTR042021005. \textit{(Equal contribution: Zheng Qin and Yabing Wang, Corresponding author: Le Wang.)}}%
	\thanks{Zheng Qin, Yabing Wang, Le Wang and Sanping Zhou are with the National Key Laboratory of Human-Machine Hybrid Augmented Intelligence, National Engineering Research Center for Visual Information and Applications, and Institute of Artificial Intelligence and Robotics, Xi'an Jiaotong University, Xi'an, Shaanxi 710049, China. (e-mail: qinzheng@stu.xjtu.edu.cn; wyb7wyb7@163.com; \{spzhou, lewang\}@mail.xjtu.edu.cn)}
	\thanks{Minghui Yang and Ming Yang are with Ant Group, Hangzhou, Zhejiang 310000, China. (e-mail: minghui.ymh@antgroup.com; m-yang4@u.northwestern.edu.) }}

\markboth{IEEE Transactions on Circuits and Systems for Video Technology,~Vol.~x, No.~x}
{Shell \MakeLowercase{\textit{et al.}}: Bare Demo of IEEEtran.cls for IEEE Journals}
%



\maketitle

\begin{abstract}
 Generating 3D human motions from text is a challenging yet valuable task. The key aspects of this task are ensuring text-motion consistency and achieving generation diversity. Although recent advancements have enabled the generation of precise and high-quality human motions from text, achieving diversity in the generated motions remains a significant challenge.  In this paper, we aim to overcome the above challenge by designing a simple yet effective text-to-motion generation method, \textit{i.e.}, Diverse-T2M. Our method introduces uncertainty into the generation process, enabling the generation of highly diverse motions while preserving the semantic consistency of the text. Specifically, we propose a novel perspective that utilizes noise signals as carriers of diversity information in transformer-based methods, facilitating a explicit modeling of uncertainty. Moreover, we construct a latent space where text is projected into a continuous representation, instead of a rigid one-to-one mapping, and integrate a latent space sampler to introduce stochastic sampling into the generation process, thereby enhancing the diversity and uncertainty of the outputs. Our results on text-to-motion generation benchmark datasets~(HumanML3D and KIT-ML) demonstrate that our method significantly enhances diversity while maintaining state-of-the-art performance in text consistency.
\end{abstract}


%
\IEEEpeerreviewmaketitle

\section{Introduction}\label{sec:introduction}
Human motion generation has extensive practical applications, such as gaming, film production, and AR/VR content creation. Generative models for human motion synthesis have been developed to automate the labor-intensive and time-consuming process of character animation, with the aim of reducing costs and enhancing development efficiency. 
Recently, there have been notable advancements in human motion generation with various conditional inputs, such as music~\cite{lee2019dancing, li2022danceformer, li2021ai}, control signals~\cite{peng2021amp, starke2022deepphase, yu2025spatio}, action categories~\cite{petrovich2021action, guo2020action2motion}, and natural language descriptions~\cite{petrovich2022temos, zeng2025progressive, ahuja2019language2pose, guo2022tm2t, kim2023flame}.
These developments offer a more intuitive and user-friendly method for animating virtual characters.
Among all conditional modalities, text-based human motion generation~(text-to-motion generation, T2M) has become a dominant focus of research, as linguistic descriptors provide a convenient and natural interface for human-computer interaction.

\begin{figure}[t]
	\begin{center}
		\includegraphics[width=1\linewidth]{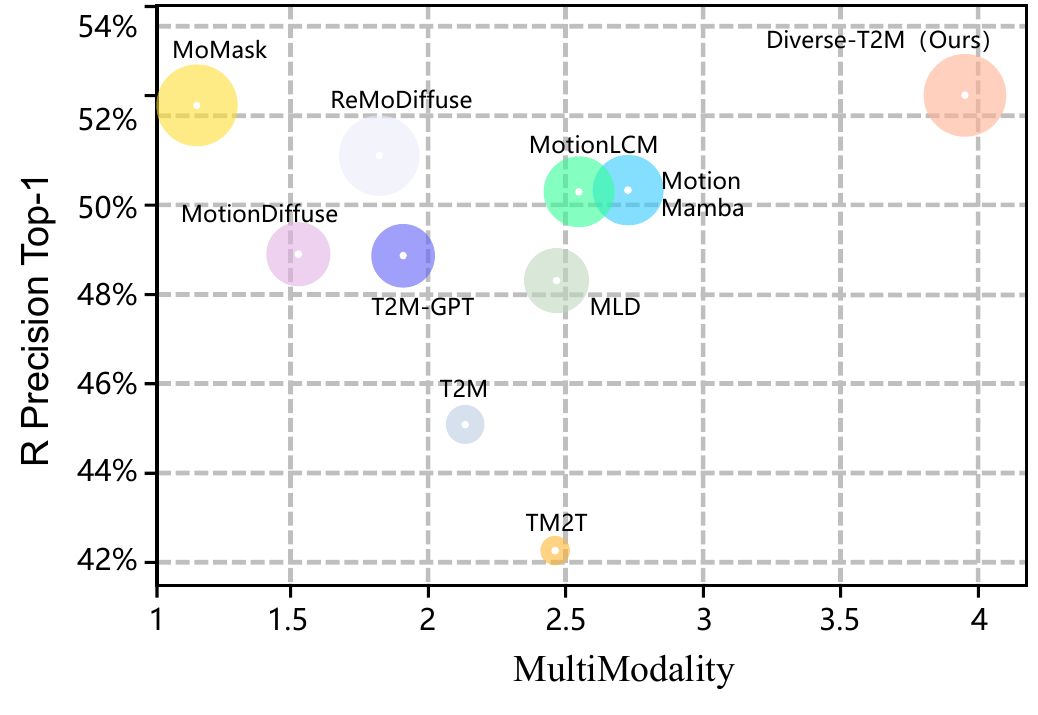}
	\end{center}
	\caption{MultiModality-R\_Precision-Fid comparisons of different text-to-motion  generation methods on HumanML3D dataset. The horizontal axis is MultiModality~(diversity), the vertical axis is R Precision Top-1~(semantic consistency), and the radius of circle reflects Fid~(fidelity). Our Diverse-T2M achieves 3.976 MultiModality, 52.5\% R Precision Top-1, 0.057 FID, outperforming all previous methods. Derails are given in Table~\ref{tab:table1}. }
	\label{fig:intro1}
\end{figure}

Existing works in this field mainly follow either the two-stage~\cite{guo2024momask, zhang2023generating} or diffusion-based~\cite{zhang2022motiondiffuse, zhang2023remodiffuse} paradigm. The former two-stage method involves:
(i) conduct motion discrete representation, \textit{e.g.}, T2M-GPT~\cite{zhang2023generating} and Att-T2M~\cite{zhong2023attt2m} utilize Vector Quantized Variational Autoencoders~(VQ-VAE)~\cite{van2017neural}, while
MoMask~\cite{guo2024momask} and T2M-HiFiGPT~\cite{wang2023t2m} introduce variants,\textit{i.e.}, residual VQVAE~\cite{zeghidour2021soundstream, defossez2022high};
\begin{figure*}[t]
	\begin{center}
		\includegraphics[width=0.9\linewidth]{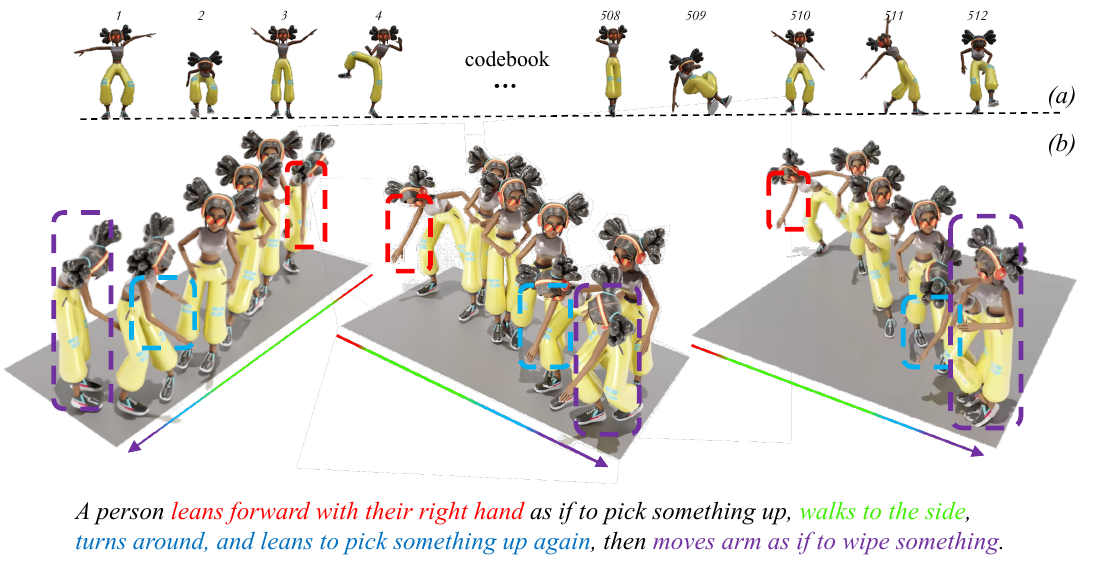}
	\end{center}
	\caption{(a) Visualization of codebook. We visualize each code feature in the codebook using the decoder. Each code comprises a spatial displacement and a fixed pose, and we specifically visualize the pose. (b) Visual results on HumanML3D. Our Diverse-T2M, when provided with a long and complex text input, generates high-quality 3D motion with great diversity diversity and precise control, \textit{i.e.}, each generated motions exactly renders the ``pick-walk-pick-wipe'' action flow~(colorful timeline), precise control of ``right hand for the first pick''~(red box), diverse action for the direction of walking, hand for second pick~(blue box), and the posture (standing, leaning, bending) when wiping~(purple box).}
	\label{fig:intro2}
\end{figure*}
(ii)) predict text-to-motion sequence, \textit{e.g.}, generative pre-trained transformers~(GPT)~\cite{radford2018improving} and mask generative transformers~\cite{chang2023muse} are employed for text-to-code sequence prediction and then decoded into motion sequences based on the codebook and decoder from the first stage.
Additionally, given the remarkable success of diffusion-based generative models in other domains~\cite{xu2022geodiff, rombach2022high}, the latter paradigm employs conditional diffusion models for human motion generation, including MDM~\cite{tevet2022motionclip}, MotionDiffuse~\cite{zhang2022motiondiffuse}, ReMoDiffuse~\cite{zhang2023remodiffuse}, FineMoGen~\cite{zhang2024finemogen}, and MLD~\cite{chen2023executing}.
However, diffusion-based paradigm directly learn text-driven generation in more complex pose space, which is challenging. In complex pose spaces, data distribution becomes sparse, making it difficult for models to capture the underlying structure of the data.
This complexity hinders the learning of the correspondence between text and motions and the slight changes may lead to visible defects in generated motion. 
Considering fidelity, training efficiency, the two-stage methods to discrete representation of motion before learning to generation ability receives more attention.

To effectively apply text-driven motion generation, 
two key requirements need to be met: \textit{text-motion consistency} and \textit{generation diversity}. The text provided by the user does not describe all the details in the motion.
For the more loosely controlled parts, there should be diverse and reasonable generation results, while maintaining consistency of textual description. 
For example, as highlighted by the purple boxes in Figure~\ref{fig:intro2} (b), there are multiple plausible postures that are consistent with the textual instruction ``moves arm as if to wipe something'', such as standing, leaning, or bending.
On the one hand, this can enhance the visual richness and dynamism of application scenes, preventing homogenization. On the other hand, diverse generations provides a broader array of foundational options for animators to perform secondary processing, thereby expanding creative possibilities and improving the overall quality of the animations.

However, we observe that although recent two-stage methods enable the generation of precise and high-quality human motions controlled by text, they still struggle to achieve the diversity mentioned above. In light of this, we conduct an in-depth analysis.
Specifically, we find that the ability of the two-stage approach to produce high-quality motions stems from discretizing real motions into a large set of basis motions.
These basis motions can then be selected and permuted to compose a variety of complex motion sequences. Because these basis motions are derived from real motion data, the authenticity of the decoded generated motions is preserved, avoiding visual defects such as gestures that violate common sense or distortions.
Additionally, these basis motions can be decoupled into spatial positions and body poses. As shown in Figure~\ref{fig:intro2} (a), each code in the VQ-VAE codebook represents a human pose.
Transformers are then used to learn a one-to-one mapping between text inputs and code sequences. By introducing small perturbations to the final generated result, different motion outputs can be obtained. However, while this process ensures strong semantic consistency with the text, its deterministic nature limits the diversity of generated motions. As Heraclitus famously said, ``Panta Rhei'' (everything flows), different individuals—and even the same individual at different times—exhibit distinct sequences of motions. Therefore, incorporating more \textit{uncertainty} into the generation process can enhance generation diversity, allowing for more realistic results.



In this paper, we propose a simple yet effective T2M method, \textit{i.e.}, Diverser-T2M, to generate highly diverse motions while maintaining semantic consistency of the text. 
As illustrated in Figure~\ref{fig:intro1}, our Diverser-T2M significantly enhances diversity while maintaining state-of-the-art performance in text consistency.
Our Diverser-T2M follows the two-stage paradigm, in which employ RVQ-VAE to conduct the motion discrete representation and learn the text-to-code sequence prediction with mask transformer. To enhance the generative power of the model, we introduce uncertainty to the generation process. 
Since diversity requires a carrier of additional information, we propose using noise as this carrier. 
Specifically, instead of only using text description as the condition signal, we incorporate a noise signal to inject uncertainty. Each inference leverages different noise signals to introduce diverse sources of information.
Moreover, the text-motion correspondence is not one-to-one, each text description should correspond to a broad solution space and be mapped to a distribution. To address this, we design a variational text-to-codes predictor by constructing a latent space where each text corresponds to a distribution. We also integrate a latent space sampler into our predictor, allowing for sampling from this distribution during inference, which significantly enhances diversity.
Extensive experiments on two benchmark datasets (HumanML3D and KIT-ML) demonstrate that our proposed Diverse-T2M outperforms the previous state-of-the-art methods at various metrics. It is worth noting that the Multimodality metric, which reflects diversity, is 30\% higher compared to the second place.


The main contribution of this work can be summarized as follows:
\begin{itemize}
    \item We analyze the potential reasons for the lack of diversity in existing methods, and propose a simple yet effective text-to-motion generation method, Diverse-T2M.
    Our method introduces uncertainty into the generation process, enabling the generation of highly diverse motions while preserving the semantic consistency of the text.
    \item We propose a novel perspective that utilizes noise signals as carriers of diversity information in transformer-based methods, facilitating a explicit  modeling of uncertainty.
    \item We construct a latent space that projects text into a continuous representation rather than a rigid one-to-one mapping, and incorporate a latent space sampler to introduce stochasticity into the generation process, thereby enhancing both the diversity and uncertainty of the outputs.
    \item Experimental results show that our proposed Diverser-T2M significantly enhances diversity while maintaining state-of-the-art performance in text consistency. And demonstrated commercial value and potential for non-human character-animated.
\end{itemize}

The remainder of the paper is organized as follows.
Section~\ref{sec:rel-work} briefly reviews related work on human motion generation.
Subsequently, we present the technical details of the proposed method in Section~\ref{sec:ESS}.
The experimental results are presented in Section~\ref{sec:experimentsAndDiscussions}.
Finally, we conclude this paper in Section~\ref{sec:conc}.

\section{Related Work} \label{sec:rel-work}

\begin{figure*}[t]
	\centering
	\resizebox{1.0\textwidth}{!}{
		\includegraphics{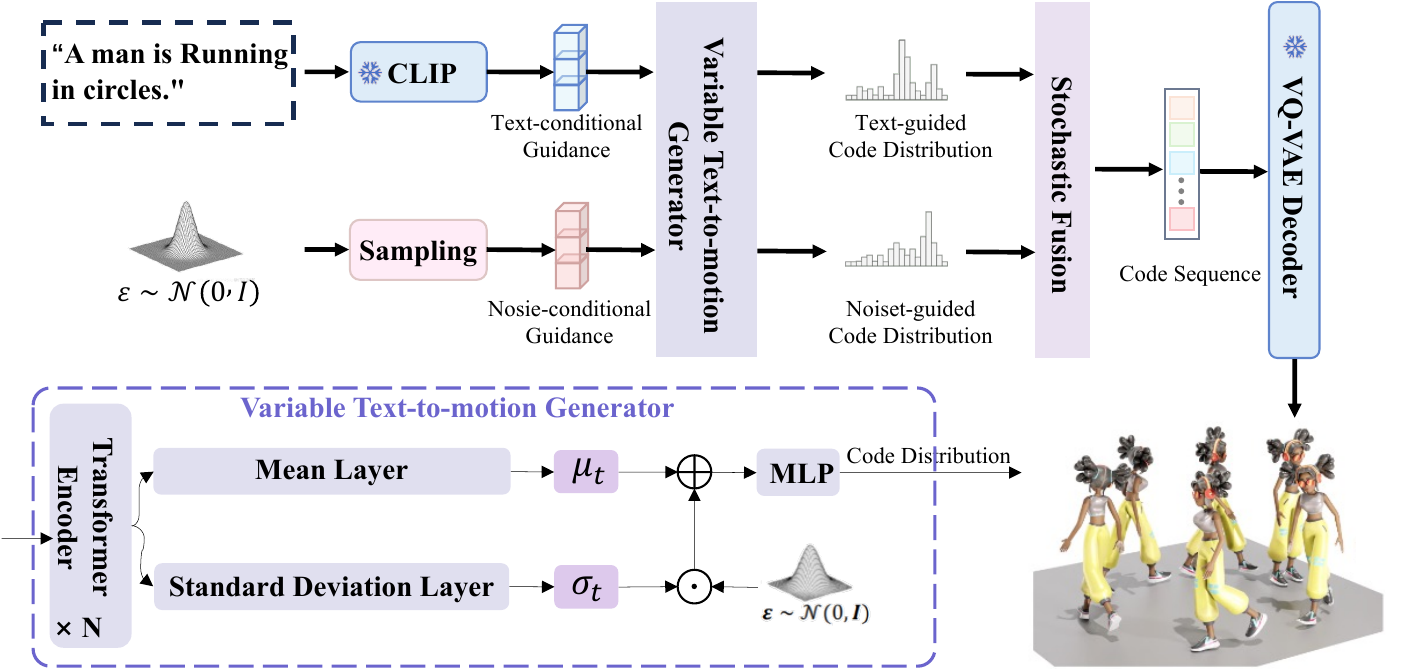} 
	}
	\caption{
    Our Diverse-T2M follows the two-stage paradigm, employing RVQ-VAE for motion discrete representation~(above) and learning text-to-code sequence prediction~(below). Rather than relying solely on text descriptions as the conditioning signal, we introduce a noise signal to inject uncertainty. Both the text and noise signals are input into the variational text-to-code predictor separately, yielding code distributions. Following a stochastic fusion process, we obtain a code sequence that aligns with the text. Finally, we utilize the codebook and decoder trained in the first stage to generate the final results.
 }
	\label{framework}
\end{figure*}

\subsection{Text-driven human motion generation.} 

Due to the user-friendly and convenient language input, text-to-motion is one of the most significant motion generation tasks. Text-to-motion is a burgeoning field focused on generating 3D human motions driven by textual descriptions. Text2Action~\cite{Ahn_Ha_Choi_Yoo_Oh_2018} utilize an RNN-based model to generate motions conditioned on brief texts. Language2Pose ~\cite{ahuja2019language2pose} further improve by learning a joint embedding space of text and pose, which is learned end-to-end using a curriculum learning approach. It emphasizes shorter and easier sequences first before moving to longer and harder ones and eventually enables the generation of motion sequences from text embeddings.
\cite{ghosh2021synthesis} propose a hierarchical two-stream sequential model to explore a finer joint-level mapping between natural language sentences and 3D pose sequences corresponding to the given motion. 
MotionCLIP~\cite{tevet2022motionclip} and ohMG~\cite{lin2023being}
model text-to-motion in an unsupervised manner using the
large pretrained CLIP~\cite{radford2021learning} model, which has been widely used in the multimodal field~\cite{wang2024dual,dong2022reading,wang2025denoising,liu2025up,qin2025versatile}. 
T2M~\cite{Guo_Zou_Zuo_Wang_Ji_Li_Cheng} extended the temporal VAE to learn the probabilistic mapping between texts and motions and it introduce a large-scale dataset HumanML3D, which is marked a significant advancement.
ACTOR~\cite{radford2021learning} design a class-agnostic transformer VAE for generating actions from predefined classes in a non-autoregressive fashion. TEMOS~\cite{Petrovich_Black_Varol_2022} added a text encoder and produced more diverse action sequences and TEACH~\cite{Athanasiou_Petrovich_Black_Varol_2022} expandeds by generating sequences of actions from multiple natural language descriptions. 
TM2T~\cite{guo2022tm2t} combines text-to-motion and motion-to-text tasks, allowing for a more comprehensive understanding of both motion and text modalities.
MotionGPT~\cite{jiang2024motiongpt} treats human motion as a foreign language and leverages language understanding and zero-shot transfer abilities of pre-trained language models.

Recently, emerging two-stage approaches and diffusion models based have dramatically changed the field of motion generation. 
Regarding the former two-stage methods, they contain two components: 1) motions are firstly discretized as tokens via vector quantization~\cite{van2017neural} and 2) then predict the code sequence of motion. 
For the first stage, recent works such as T2M-GPT~\cite{zhang2023generating} and Att-T2M~\cite{zhong2023attt2m} utilize the Vector Quantized Variational Autoencoders~(VQVAE)~\cite{van2017neural} for motion discrete representation.
Additionally, MoMask~\cite{guo2024momask} and T2M-HiFiGPT~\cite{wang2023t2m} employ variants like residual VQVAE~\cite{zeghidour2021soundstream, defossez2022high} for their approaches.
For the second stage, T2M-GPT~\cite{zhang2023generating} introduce Generative Pretrained Transformer (GPT) for text-to-code sequence prediction. MoMask~\cite{guo2024momask} improve it by employ Masked Transformer for code sequence generation conditioned by text.
Given the remarkable success of diffusion-based generative models in other domains~\cite{xu2022geodiff, rombach2022high}, the former method employ a conditional diffusion model for human motion generation by MDM~\cite{tevet2022motionclip}, MotionDiffuse~\cite{zhang2022motiondiffuse}, ReMoDiffuse~\cite{zhang2023remodiffuse}, FineMoGen~\cite{zhang2024finemogen} and MLD~\cite{chen2023executing}. This approach aims to learn a robust probabilistic mapping from textual descriptors to motion sequences. 
However, diffusion-based models often exhibit lower motion-text matching accuracy and slower inference speed due to the iterative nature of diffusion processes. Considering fidelity, training efficiency, the two-stage methods has attract more attention.
Although the two-stage approach can achieve high fidelity and semantic consistency, the generation process lacks stochasticity, leading to reduced diversity in the generated results.
In contrast, our Diverse-T2M is able to generate diverse motion sequences while maintaining a high level of semantic consistency with the accompanying text.

\begin{figure*}[t]
	\centering
	\resizebox{0.85\textwidth}{!}{
		\includegraphics{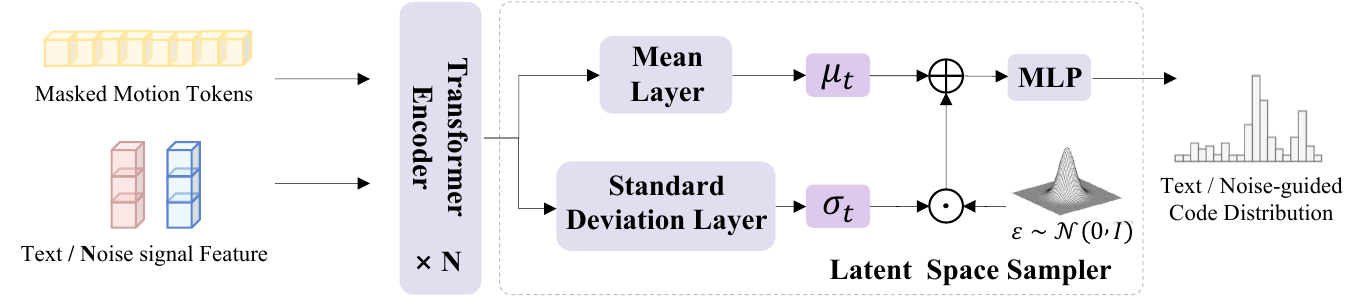} 
	}
	\caption{
 Illustration of the Variational Text-to-Code Predictor. This model takes masked motion tokens and signal feature as input, producing a code distribution that corresponds to the input signals.
 }
	\label{framework2}
\end{figure*}
    
\subsection{Motion Discrete Representation.} 
Motion data is important in the development of motion synthesis tasks. The presentation of motion data in the generative algorithm have the straightforward joint positions and the Master Motor Map~(MMM) format. For mainstream datasets, \textit{i.e.}, HumanML3d and KIT, they employ two motion representations: 1) the classical SMPL-based~\cite{Chen_Pang_Yang_Ma_Xu_Yu_2021, Kocabas_Athanasiou_Black_2020,  Loper_Mahmood_Romero_Pons-Moll_Black_2015} motion parameters and 2) the redundant hand-crafted motion feature~\cite{Guo_Zou_Zuo_Wang_Ji_Li_Cheng, Starke_Zhang_Komura_Saito_2019} with a combination of joints features. In order to generate actions with plausibility and fidelity, some work has discretized real actions into a number of representations and used the discretized representations to de-combine them into complete action sequences. Recent works such as T2M-GPT~\cite{zhang2023generating} and Att-T2M~\cite{zhong2023attt2m} utilize the Vector Quantized Variational Autoencoders~(VQVAE)~\cite{van2017neural} for motion discrete representation.
Additionally, MoMask~\cite{guo2024momask} and T2M-HiFiGPT~\cite{wang2023t2m} employ variants like residual VQVAE~\cite{zeghidour2021soundstream, defossez2022high} for their approaches. 




\section{Proposed Method}
\label{sec:ESS}    

\subsection{Overview of Diverse-T2M}
\noindent\textbf{Notation.}
Text-to-motion generation aims to generate 3D human motions from text descriptions. As shown in Figure~\ref{framework}, given the text $s$ as conditional information, it is expected to generate a motion sequence $M = \{m_t\}_{t=1}^T$ consistent with the text $s$, where $T$ denotes the motion length. Here, $m_t$ represents the pose at time $t$, which contains information about the angular and linear velocities of the body's root, as well as the positions, velocities, and rotations of the local joints in the root space. We denote the text-to-motion generation datasets by $\mathcal{X} = \{s_i, M_i)\}_{i=1}^N$, which is the collection of $N$ semantically corresponding text-motion pairs.
Our Diverse-T2M follows a simple and classic two-stage paradigm~\cite{guo2024momask,zhang2023generating,guo2022tm2t}, which involves following two stages: 
\begin{itemize}
\item {\bf Stage 1: Motion Discrete Representation.} We learn discrete representations for motion by learning the mapping between the motion space and a discretized feature space, \textit{i.e.}, a code space, where each code represents a discretized feature. In particular, it consists of three key components: the codebook $C$, which defines the discrete representation space; the encoder $\mathcal{E}$, which maps from the motion space to the discretized feature space; and the decoder $\mathcal{D}$, which performs the reverse mapping. With a well-learned discrete representation, any motion can be transformed into a sequence of code that represents discretized features with feature extraction by $\mathcal{E}$ and quantization by $C$~(Section~\ref{sec4.2}). 
\item {\bf Stage 2: Text-to-codes Predication.} 
After the discrete representation of motion, we directly focus on the process of generating code sequence in the discretized feature space conditional on text, which are then mapped to a motion sequence via the aforementioned trained decoder $\mathcal{D}$. With a discretized feature space as the medium, the text-to-motion generation process is completed.
\end{itemize}

Due to individual differences, each person's behavioral patterns are unique, and even the same individual will not exhibit exactly the same behavior repeatedly. Therefore, a given text description should correspond to various motions, creating a broad solution space for the text.
Despite some recent two-stage methods enable the generation of precise and high-quality human motions controlled by text, they still struggle to achieve generation diversity.
To address this challenge, we introduces uncertainty into the generation process, enabling the generation of highly diverse motions while preserving the semantic consistency of the text.
Specifically, we propose a novel perspective that utilizes noise signals as carriers of diversity information in transformer-based methods, \textit{i.e.}, conducts motion generation under noise and text conditions~(see Section~\ref{Motion Generation under Noise and Text Conditions}).
Moreover, we construct a latent space to map text to a data
distribution, replacing the one-to-one mapping, and integrate a latent space sampler to introduce random sampling into the generation process, further enhancing the uncertainty within the generation~(see Section~\ref{Variational Text-to-codes Predictor}).

\subsection{Diverse-T2M}
\subsubsection{Motion Discrete Representation}
\label{sec4.2}

In this section, we aim to apply the VQ-VAE~\cite{van2017neural} to learn discrete representations, which are beneficial for generative models. 
As shown in~Figure~\ref{framework}-top, VQ-VAE aims to reconstruct the motion sequence using an auto-encoder and a learnable $K$-size codebook $C = \{c_k\}_{k=1}^K$, where each code $k$ corresponds to a discretized feature $c_k \in \mathbb{R}^{d_c}$ and $d_c$ represents the discretized feature dimension. In detail, given the motion sequence $M = \{m_t\}_{t=1}^T$, the encoder $\mathcal{E}$ of the autoencoder encodes it into a latent feature sequence denoted as $Z = \{z_i\}_{i=1}^{T/l}$ where $z_i \in \mathbb{R}^{d_c}$ and $l$ is the temporal downsampling rate of $\mathcal{E}$. For each latent feature $z_i$, the quantization through $C$ is to find the most similar element in $C$ and form $\hat{Z} = \{\hat{z}_i\}_{i=1}^{T/l}$ as follows:
\begin{equation}
\hat{z}_i=\underset{c_k \in  C}{\arg \min }\left\|z_i-c_k\right\|_2,
\end{equation}
where $\hat{z}_i$ represents the quantized feature of $z_i$. Finally, the quantized feature sequence $\hat{Z}$ is then decoded back into motion sequence $M$ by the decoder $\mathcal{D}$.

To reduce information loss and improve reconstruction quality, we implement residual quantization in motion reconstruction following~\cite{zeghidour2021soundstream}. The objective is to approximate each feature $\bm{z}_i$ as closely as possible with $\hat{z}_i$ by modeling the residuals between them, \textit{i.e.}, $r_i = z_i - \hat{z}_i$. Concretely, we set $V$ layer residual quantization with $V$ codebooks $\{C^v\}_{v=0}^V$. We recursively conduct residual quantization as follows:
\begin{equation}
\begin{aligned}
\hat{r}_i^v =\underset{c_k^v \in C^v}{\arg \min }\left\|r_i^v-c_k^v\right\|_2, r_i^0 = z_i\\
\quad r_i^{v+1}=r_i^v-\hat{r}_i^v, \ v = 0, 1, \ldots, V.
\end{aligned}
\end{equation}
where $r_i^v$ and $\hat{r}_i^v$ represent the $v$-th layer's residual feature and quantized residual feature respectively.
After residual quantization, the final approximation of latent sequence $z_i$ is the sum of all quantized sequences $\sum_{v=0}^V \hat{r}^v_i$, which is then fed into decoder $\mathcal{D}$ for motion reconstruction. 




\subsubsection{Motion Generation under Noise and Text Conditions}
\label{Motion Generation under Noise and Text Conditions}


Distinguishing from motion generation conditioned only on text, we include noise as one of the conditions for motion generation as well, increasing the uncertainty of the generation, and thus improving the diversity.

Particularly, instead of training a text-conditional model $p_\theta(z | s)$ alone, we simultaneously train a noise-conditional model $p_\theta(z | \epsilon)$ alongside it. 
A variational text-to-codes predictor (Section~\ref{Variational Text-to-codes Predictor}) is used to parameterize both models, $p_\theta(z | s)$ and $p_\theta(z | \epsilon)$. Similar to the classifier-free diffusion guidance approach, while training separate models for $p_\theta(z | s)$ and $p_\theta(z | \epsilon)$ is possible, joint training is simpler to implement, does not complicate the training process, and avoids increasing the total number of parameters. The noise-conditional and text-conditional models are trained jointly by fusion the text-conditional signal $s$ with the noise-conditional signal with a certain probability $p_\text{noise}$, which is treated as a hyperparameter.

Specifically, given the text $s$, we use CLIP to extract the text feature as text signal $g_s$. We sample a noise signal $g_\epsilon$ from Gaussian noise that is the same size as the text signal. The text signal and noise signal are then fed into the variational text-to-motion predictor separately to obtain codes probability distribution $I_\text{text}, I_\text{noise} \in \mathbb{R}^{T \times K}$, where $T$ and $K$ represent the predicted motion length and size of the codebook respectively. At timestamp $t$, the probability distribution of the motion codes is as follows:
\begin{equation}
\begin{aligned}
P(m_t=c_k) = p_k^t ,  \ \  k = 1,...,K. \\
\end{aligned}
\end{equation}

Similar to classifier-free diffusion guidance, we fuse the probability distributions obtained from the text signal and the noise signal as follows:
\begin{equation}
\begin{aligned}
I = (1+w)I_\text{text} - wI_\text{noise},  \\
\end{aligned}
\end{equation}
$w$ is the fusion weight, which can balance the diversity and semantic consistency with text description.
Then, we select the code sequence $\{c_i\}_{i=1}^{T/l}$ that matches the text signal through the probability distribution $I$. 

Once the code sequence consistent with the given text description is obtained, it is translated into the motion feature sequence $\{f_i\}_{i=1}^{T/l}$ using the codebook. Finally, a well-trained decoder converts the motion feature sequence into the motion sequence $\hat{M} = \{m_t\}_{t=1}^T$.

\noindent\textbf{Length Stochastic Enhancement.} 
The text-to-motion generation datasets $\mathcal{X} = \{s_i, M_i)\}_{i=1}^N$ collects $N$ semantically corresponding text-motion pairs. 
Although text and motion are semantically corresponding in each pair, the motion in each pair must not be the unique solution to text. Each text has many motions corresponding to it, and this is also true in terms of the motion length. 
To increase data diversity during training, we randomly stretch and compress the motion sequences in equal proportions.
Considering that some text includes speed descriptions such as 'fast' and 'slow,' we avoid data conflicts from data augmentation by only enhancing the length of motions corresponding to noisy signals during training. This approach maintains semantic consistency while enhancing data diversity.

\subsection{Variational Text-to-codes Predictor}
\label{Variational Text-to-codes Predictor}
The input to the variational text-to-codes predictor is the signal feature $g$, which can be either a text signal $g_s$ or a noise signal $g_\epsilon$. The output is the codes probability distribution, specifically $I_\text{text}$ for $g_s$ and $I_\text{noise}$ for $g_\epsilon$, both belong to $\mathbb{R}^{T \times K}$.

For the traditional transformer-based generative models in two-stage methods~\cite{guo2024momask, zhang2023generating}, it construct one-to-one ``text-to-codes'' prediction for the $(g, I)$ pairs as follows:
\begin{equation}
\begin{aligned}
I &= f_P(g), \\
\end{aligned}
\end{equation}
where $f_P$ is implemented with the Generative Pretrained Transformer or mask transformer. We use the mask transformer here.


To encourage greater diversity, we integrate latent space sampler within our variational text-to-codes predictor, aimed at enhancing expressive diversity, as shown in Figure~\ref{framework2}.
In details, we construct a latent space and map signal feature into this latent space. Within this latent space, each signal feature corresponds to a data distribution rather than a fixed vector, as follows:
\begin{equation}
\begin{aligned}
&z \sim p(z \mid g).
\end{aligned}
\end{equation}

Specifically, as shown in~\ref{framework2}, for a given signal feature $g$ during training, we concat it with its corresponding masked motion tockens and input into $N$ transformer layers. Then obtain the fusion feature $e$. Then latent space sampler predicts a distribution $p(z \mid e)$ based on $e$, from which a latent embedding $z$ is sampled. Next, the latent embedding $z$ is passed through a multi-layer perceptron (MLP) with residual connections, mapping it to code probability distribution through the codebook. During infernce, the masked motion tockens are replaced with an empty tokens.

Then, we approximate the posterior probability model $p(z \mid e)$ with a parametric inference model $q_\phi(z \mid e)$. With $\phi$ indicating the parameters of this inference model, also called the variational parameters. We optimize the variational parameters $\phi$ such that:
\begin{equation}
\begin{aligned}
q_\phi(z \mid e) &\approx p(z \mid e). \\
\end{aligned}
\end{equation}
The distribution $q_\phi(z \mid e)$ is parameterized using deep neural networks:
\begin{equation}
\begin{aligned}
&(\mu, log\sigma^2) = EncoderNet(e), \\
&q_\phi(z \mid e)=\mathcal{N}\left(z ; \mu, diag\left(\sigma^2\right)\right),
\end{aligned}
\end{equation}
we assume that $z$ follows a multivariate Gaussian distribution where each dimension is independent. We implement the mean layer and standard deviation layer with self attention layer to predicts a mean vector $\mu$ and a log magnitude variance vector $log\sigma^2$ of the multivariate Gaussian distribution based on $e$. By leveraging the reparameterization technique, a $z$ is sampled as:
\begin{equation}
\begin{aligned}
z = \mu+\sigma \odot \epsilon , \quad \epsilon  \sim \mathcal{N}(0, I), \\
\end{aligned}
\end{equation}
Note that latent space sampler is differentiable with the reparameterization technique.

\subsection{Training Diverse-T2M}
\subsubsection{Optimization Goal of Motion Discrete Representation}
Overall, the residual VQ-VAE is trained via a motion reconstruction loss combined with a latent embedding loss at each quantization layer:
\begin{equation}
\begin{aligned}
\mathcal{L}_{m d r}=\|M-\hat{M}\|_1+\beta \sum_{v=1}^V\left\|R^v-\operatorname{sg}\left[\hat{R}^v\right]\right\|_2^2
\end{aligned}
\end{equation}
where $\operatorname{sg}[\cdot]$ denotes the stop-gradient operation, and $\beta$ a weighting factor for embedding constraint. The codebooks are updated via exponential moving average and codebook reset following~\cite{guo2024momask, zhang2023generating}.

\subsubsection{Optimization Goal of Motion Generation under Noise and Text Conditions}
We jointly train the noise-conditional model and the text-conditional model. Specifically we mix the training samples of noise signal and text signal following a certain ratio to train the model.
The loss function for supervising model learning are as follows:
\begin{equation}
\mathcal{L} = \mathcal{L}_{\text {tran }} + \mathcal{L}_{\text {kl}},
\end{equation}
$\mathcal{L}_{\text {tran }}$ is used to supervise the mask transformer to generate the correct probability distribution, and $\mathcal{L}_{\text {kl}}$ is used to supervise the latent space to learn a distribution closer to a Gaussian distribution.

For loss function $\mathcal{L}_{\text {tran}}$, 
each signal-code sequence pair is a training sample. Our training goal is to predict the masked tockens. 
We directly maximize the log-likelihood of the data distribution $p(\hat{M} \mid s)$:
\begin{equation}
\begin{aligned}
\mathcal{L}_{\text {tran }}=\mathbb{E}_{\hat{M} \sim p(\hat{M})}[-\log p(\hat{M} \mid s)],
\end{aligned}
\end{equation}
the likelihood of the full sequence is denoted as follows: 
\begin{equation}
\begin{aligned}
p(\hat{M} \mid s)=\prod_{i=1}^{T/l} \left(p\left(\hat{m}_t \mid s\right) \cdot \left(1 - [mask]_t\right) + [mask]_t\right),
\end{aligned}
\end{equation}
where $[mask]_t$ indicates whether the $i$-th token is masked, which is set to 0 if masked, and 1 otherwise. The residual layer is predicted without using mask following~\cite{guo2024momask}.

For supervising the latent space to learn a distribution closer to a Gaussian distribution, $\mathcal{L}_{\text {kl}}$ is difined as:
\begin{equation}
\mathcal{L}_{\text {kl}}=\frac{1}{D} \sum_{i=1}^D K L\left(\mathcal{N}(\mu_i,\sigma_i), \mathcal{N}(0,1)\right),
\end{equation}
$D$ represents the feature dimension and $KL$ represents the Kullback-Leibler divergence.

\section{Experiments}
\label{sec:experimentsAndDiscussions}

\begin{table*}[t]
	\centering
	\caption{Comparison with the state-of-the-art methods on HumanML3D~\cite{human3d} test set. We compute standard metrics following Guo et al.~\cite{human3d}. For each metric, we repeat the evaluation 20 times and report the average with 95\% confidence interval. § reports results using ground-truth motion length. Bold face indicates the best result, while blue refers to the second best.}
    \resizebox{1\linewidth}{!}{
\begin{tabular}{l|c|ccccc}
\hline
\multirow{2}{*}{\textbf{Methods}}                & \multirow{2}{*}{\textbf{MultiModality$\uparrow$}} & \multicolumn{3}{c}{\textbf{R Precision$\uparrow$}}                                                                       & \multirow{2}{*}{\textbf{FID$\downarrow$}} & \multirow{2}{*}{\textbf{MultiModal Dist$\downarrow$}} \\ \cline{3-5}
                                                 &                                                   & \textbf{Top 1}                         & \textbf{Top 2}                         & \textbf{Top 3}                         &                                           &                                                       \\ \hline
Seq2Seq~\cite{lin2018generating}                                          & -                                                 & \et{0.180}{.002}                       & \et{0.300}{.002}                       & \et{0.396}{.002}                       & \et{11.75}{.035}                          & \et{5.592}{.007}                                      \\
Language2Pose~\cite{ahuja2019language2pose}                                    & -                                                 & \et{0.246}{.002}                       & \et{0.387}{.002}                       & \et{0.486}{.002}                       & \et{11.02}{.046}                          & \et{5.296}{.008}                                      \\
Text2Gesture~\cite{bhattacharya2021text2gestures}                                     & -                                                 & \et{0.165}{.001}                       & \et{0.267}{.002}                       & \et{0.345}{.002}                       & \et{5.012}{.030}                          & \et{6.030}{.008}                                      \\
Hier~\cite{ghosh2021synthesis}                                             & -                                                 & \et{0.301}{.002}                       & \et{0.425}{.002}                       & \et{0.552}{.004}                       & \et{6.532}{.024}                          & \et{5.012}{.018}                                      \\
MoCoGAN~\cite{tulyakov2018mocogan}                                          & \et{0.019}{.000}                                  & \et{0.037}{.000}                       & \et{0.072}{.001}                       & \et{0.106}{.001}                       & \et{94.41}{.021}                          & \et{9.643}{.006}                                      \\
Dance2Music~\cite{lee2019dancing}                                      & \et{0.043}{.001}                                  & \et{0.033}{.000}                       & \et{0.065}{.001}                       & \et{0.097}{.001}                       & \et{66.98}{.016}                          & \et{8.116}{.006}                                      \\
TEMOS§                                           & \et{0.038}{.018}                                  & \et{0.424}{.002}                       & \et{0.612}{.002}                       & \et{0.722}{.002}                       & \et{3.734}{.028}                          & \et{3.703}{.008}                                      \\
TM2T~\cite{guo2022tm2t}                                             & \et{2.424}{.093}                                  & \et{0.424}{.003}                       & \et{0.618}{.003}                       & \et{0.729}{.002}                       & \et{1.501}{.017}                          & \et{3.467}{.011}                                      \\
T2M~\cite{Guo_Zou_Zuo_Wang_Ji_Li_Cheng}                     & \et{2.219}{.074}                                  & \et{0.455}{.003}                       & \et{0.636}{.003}                       & \et{0.736}{.002}                       & \et{1.087}{.021}                          & \et{3.347}{.008}                                      \\
MDM~\cite{shafir2023human}                        & \ets{2.799}{.072}                                 & -                                      & -                                      & \et{0.611}{.007}                       & \et{0.544}{.044}                          & \et{5.566}{.027}                                      \\
MLD~\cite{chen2023executing}                     & \et{2.413}{.079}                                  & \et{0.481}{.003}                       & \et{0.673}{.003}                       & \et{0.772}{.002}                       & \et{0.473}{.013}                          & \et{3.196}{.010}                                      \\
MotionDiffuse~\cite{zhang2022motiondiffuse}      & \et{1.553}{.042}                                  & \et{0.491}{.001}                       & \et{0.681}{.001}                       & \et{0.782}{.001}                       & \et{0.630}{.001}                          & \et{3.113}{.001}                                      \\
T2M-GPT~\cite{zhang2023generating}               & \et{1.831}{.048}                                  & \et{0.492}{.003}                       & \et{0.679}{.002}                       & \et{0.775}{.002}                       & \et{0.141}{.005}                          & \et{3.121}{.009}                                      \\
ReMoDiffuse~\cite{zhang2023remodiffuse}          & \et{1.795}{.043}                                  & \et{0.510}{.005}                       & \et{0.698}{.006}                       & \et{0.795}{.004}                       & \et{0.103}{.004}                          & \et{2.974}{.016}                                      \\
MoMask~\cite{guo2024momask} & \et{1.241}{.040}             &  \ets{0.521}{.002} & \ets{0.713}{.002} &  \ets{0.807}{.002} &  \etb{0.045}{.002}    & \ets{2.958}{.008}                \\
MotionLCM~\cite{dai2024motionlcm}                                        & \et{2.259}{.092}                                  & \et{0.502}{.003}                       & \et{0.698}{.002}                       & \et{0.798}{.002}                       & \et{0.304}{.012}                          & \et{3.012}{.007}                                      \\
Motion Mamba~\cite{zhang2025motion}                                     & \et{2.294}{.058}                                  & \et{0.502}{.003}                       & \et{0.693}{.002}                       & \et{0.792}{.002}                       & \et{0.281}{.009}                          & \et{3.060}{.058}                                      \\ \hline
Ours                                             & \etb{3.976}{.155}                                 & \etb{0.525}{.002}                      & \etb{0.717}{.003}                      & \etb{0.809}{.002}                      & \ets{0.057}{.003}                         & \etb{2.941}{.010}                                     \\ \hline
\end{tabular}
		}
	\label{tab:table1}
\end{table*}

\begin{table*}[t]
	\centering
	\caption{Comparison with the state-of-the-art methods on KIT-ML~\cite{plappert2016kit} test set. We compute standard metrics following Guo et al.~\cite{human3d}. For each metric, we repeat the evaluation 20 times and report the average with 95\% confidence interval. § reports results using ground-truth motion length. Bold face indicates the best result, while blue refers to the second best.}
    \resizebox{1\linewidth}{!}{
\begin{tabular}{l|c|ccccc}
\hline
                                            &                         & \multicolumn{3}{c}{R Precision$\uparrow$}                 &                   &                             \\ \cline{3-5}
Methods                                     & MultiModality$\uparrow$ & Top 1             & Top 2             & Top 3             & FID$\downarrow$   & MultiModal Dist$\downarrow$ \\ \hline
Seq2Seq~\cite{lin2018generating}                                     & -                       & \et{0.103}{.003}  & \et{0.178}{.005}  & \et{0.241}{.006}  & \et{24.86}{.348}  & \et{7.960}{.031}            \\
Language2Pose~\cite{ahuja2019language2pose}                               & -                       & \et{0.221}{.005}  & \et{0.373}{.005}  & \et{0.483}{.005}  & \et{6.545}{.072}  & \et{5.147}{.030}            \\
Text2Gesture~\cite{bhattacharya2021text2gestures}                                & -                       & \et{0.178}{.005}  & \et{0.255}{.004}  & \et{0.338}{.005}  & \et{12.12}{.183}  & \et{6.964}{.029}            \\
Hier~\cite{ghosh2021synthesis}                                        & -                       & \et{0.255}{.006}  & \et{0.432}{.007}  & \et{0.531}{.007}  & \et{5.203}{.107}  & \et{4.986}{.027}            \\
MoCoGAN~\cite{tulyakov2018mocogan}                                     & \et{0.250}{.009}       & \et{0.022}{.002}  & \et{0.042}{.003}  & \et{0.063}{.003}  & \et{82.69}{.242}  & \et{10.47}{.012}            \\
Dance2Music~\cite{lee2019dancing}                                 & \et{0.062}{.002}       & \et{0.031}{.002}  & \et{0.058}{.002}  & \et{0.086}{.003}  & \et{115.4}{.240}  & \et{10.40}{.016}            \\
TEMOS§                                      & \et{0.532}{.018}       & \et{0.353}{.002}  & \et{0.561}{.002}  & \et{0.687}{.002}  & \et{3.717}{.028}  & \et{3.417}{.008}            \\
TM2T~\cite{guo2022tm2t}                     & \ets{3.292}{.081}       & \et{0.280}{.005}  & \et{0.463}{.006}  & \et{0.587}{.005}  & \et{3.599}{.153}  & \et{4.591}{.026}            \\
T2M~\cite{Guo_Zou_Zuo_Wang_Ji_Li_Cheng}                & \et{2.052}{.107}        & \et{0.361}{.005}  & \et{0.559}{.007}  & \et{0.681}{.007}  & \et{3.022}{.107}  & \et{3.488}{028}             \\
MDM~\cite{shafir2023human}                   & \et{1.907}{.214}        & -                 & -                 & \et{0.396}{.004}  & \et{0.497}{.021}  & \et{9.191}{.022}            \\
MLD~\cite{chen2023executing}                & \et{2.192}{.071}       & \et{0.390}{.008}  & \et{0.609}{.008}  & \et{0.734}{.007}  & \et{0.404}{.027}  & \et{3.204}{.027}            \\
MotionDiffuse~\cite{zhang2022motiondiffuse} & \et{0.730}{.013}        & \et{0.417}{.004}  & \et{0.621}{.004}  & \et{0.739}{.004}  & \et{1.954}{.062}  & \et{2.958}{.005}            \\
T2M-GPT~\cite{zhang2023generating}                & \et{1.570}{.039}        & \et{0.416}{.006}  & \et{0.627}{.006}  & \et{0.745}{.006}  & \et{0.514}{.029}  & \et{3.007}{.023}            \\
ReMoDiffuse~\cite{zhang2023remodiffuse}     & \et{1.239}{.028}        & \et{0.427}{.014} & \et{0.641}{.004} & \et{0.765}{.055} & \etb{0.155}{.006} & \et{2.814}{.012}           \\
MoMask~\cite{guo2024momask}                                      & \et{1.131}{.043}        & \ets{0.433}{.007} & \etb{0.656}{.005} & \etb{0.781}{.005} & \ets{0.204}{.011} & \etb{2.779}{.022}           \\ 
Motion Mamba~\cite{zhang2025motion}                                  & \et{1.678}{.064}        & \et{0.419}{.006} & \et{0.645}{.005} & \et{0.765}{.006} & \et{0.307}{.041} & \et{3.021}{.025}           \\ \hline
Ours                                                  & \etb{3.462}{.122}        & \etb{0.437}{.005} & \ets{0.654}{.005} & \ets{0.774}{.005} & \et{0.257}{.013} & \ets{2.801}{.024}           \\ \hline
\end{tabular}
		}
	\label{tab:table2}
\end{table*}

In this section, we evaluate the text-to-motion generation performance of our proposed Diverse-T2M and carry out detailed ablation studies to explore the contribution of each component to the performance of Diverse-T2M. Meanwhile, we compare our method with existing state-of-the-art methods on three two benchmark datasets.

\begin{table*}[t]
\normalsize
\centering
\caption{
\textcolor{black}{Ablation study about variable text-to-codes predictor~(VP) and the noise signal~(NS) on HumanML3D test set. We compute standard metrics following Guo et al.~\cite{human3d}. For each metric, we
repeat the evaluation 20 times and report the average with 95\% confidence interval.  }}
\label{abeach}
\setlength{\tabcolsep}{1.0em}%
\begin{tabular}{lcccccc}
\hline
               &                         & \multicolumn{3}{c}{R Precision$\uparrow$}              &                  &                             \\ \cline{3-5}
Methods        & MultiModality$\uparrow$ & Top 1            & Top 2            & Top 3            & FID$\downarrow$  & MultiModal Dist$\downarrow$ \\ \hline
Baseline~(VQ)  & \et{1.130}{.043}        & \et{0.505}{.002} & \et{0.698}{.002} & \et{0.794}{.003} & \et{0.089}{.005} & \et{3.043}{.008}            \\
+ VP           & \et{1.904}{.060}        & \et{0.509}{.002} & \et{0.700}{.003} & \et{0.795}{.002} & \et{0.085}{.003} & \et{3.040}{.009}            \\
+ NS           & \et{4.047}{.146}        & \et{0.510}{.003} & \et{0.702}{.002} & \et{0.797}{.002} & \et{0.083}{.003} & \et{3.031}{.010}            \\ \hline
Baseline~(RVQ) & \et{1.221}{.040}        & \et{0.518}{.002} & \et{0.714}{.002} & \et{0.807}{.002} & \et{0.058}{.002} & \et{2.962}{.008}            \\
+ VP           & \et{1.891}{.055}        & \et{0.524}{.002} & \et{0.716}{.002} & \et{0.810}{.002} & \et{0.054}{.002} & \et{2.952}{.008}            \\
+ NS           & \et{3.976}{.155}         & \et{0.525}{.002} & \et{0.717}{.003} & \et{0.809}{.002} & \et{0.057}{.003} & \et{2.941}{.010}            \\ \hline
\end{tabular}
\label{table3}
\end{table*}

\begin{figure*}[t]
\centering
\includegraphics[width=1.0\textwidth]{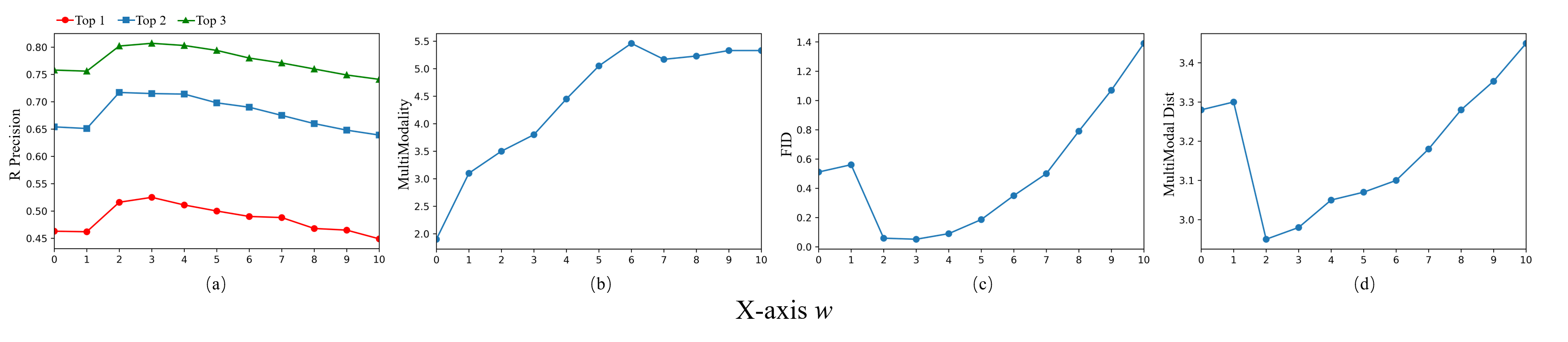}
\caption{
Evaluation sweep over the mixing ratio $w$ for the results of the noise signal and text signal during inference, assessing metrics such as R Precision, MultiModality, FID, and MultiModal Dist. The larger the value of $w$, the greater the effect of the noise signal on the generated results.
}
\label{figure5}
\end{figure*}

\subsection{Experimental Setting}

\textbf{Evaluation Datasets}. 
In this study, we evaluate our proposed method on two widely recognized text-to-motion generation datasets.  KIT Motion-Language (KIT-ML)~\cite{plappert2016kit} and  HumanML3D~\cite{Guo_Zou_Zuo_Wang_Ji_Li_Cheng}.  We follow the evaluation protocol proposed in HumanML3D~\cite{Guo_Zou_Zuo_Wang_Ji_Li_Cheng, guo2022tm2t}.

\noindent\textbf{Evaluation Metrics}. 
We follow the common metrics of prior works~\cite{Guo_Zou_Zuo_Wang_Ji_Li_Cheng} to evaluate the text-to-motion generation performance. Global representations of motion and text descriptions are first extracted with the pre-trained network in~\cite{Guo_Zou_Zuo_Wang_Ji_Li_Cheng}, and then measured by the following five metrics:

\begin{itemize}
	\item R-Precision. Given one motion sequence and 32 text descriptions (1 ground-truth and 31 randomly selected mismatched descriptions), we rank the Euclidean distances between the motion and text embeddings. Top-1, Top-2, and Top-3 accuracy of motion-to-text retrieval are reported.
	\item Frechet Inception Distance (FID). We calculate the distribution distance between the generated and real motion using FID~\cite{heusel2017gans} on the extracted motion features.
        \item Multimodal Distance (MM-Dist). The average Euclidean distances between each text feature and the generated motion feature from this text.
        \item Multimodality (MModality). For one text description, we generate 30 motion sequences forming 10 pairs of motion. We extract motion features and compute the average Euclidean distances of the pairs. We finally report the average over all the text descriptions.
\end{itemize}

\noindent\textbf{Experimental Settings}.
We implemented our Diverse-T2M in PyTorch~\cite{paszke2019pytorch} by performing all experiments on 1×3090 GPUs.
In our experiments, the downsampling rate of encoder and the residual lay number are set to 4 and 6 respectively. The size of codebook is 512 × 512, \textit{i.e.}, 512 512-dimension dictionary vectors. Following~\cite{human3d,guo2024momask,zhang2023generating}, the dataset HumanML3D and KIT-ML are extracted into motion features with dimensions 263 and 251 respectively, which related to local joints position, velocity, and rotations in root space as well as global translation and rotations. The features are calculated from  SMPL~\cite{Loper_Mahmood_Romero_Pons-Moll_Black_2015} joints. And the joint number for these two datasets is set to 22 and 21. During inference, the mixing ratio $w$ for the generation of the noise signal and text signal results is set to 3. During training, the KL divergence is multiplied by 1e-5 when supervised. 

\subsection{Benchmark Evaluation}
This section will compare our method with state-of-the-art text-to-motion generation methods on HumanML3D~\cite{Guo_Zou_Zuo_Wang_Ji_Li_Cheng} and KIT Motion-Language (KIT-ML)~\cite{plappert2016kit}.
The results are presented in Table~\ref{tab:table1} and \ref{tab:table2}, respectively.
They show that our method achieves the state-of-the-art performance, state-of-the-art, which substantially increase the diversity of generation while ensuring that the generated motion sequences have a high semantic consistency with the conditional text, validating the efficacy of our proposed method. 

\textbf{HumanML3D}. 
Table~\ref{tab:table1} presents the comparison results of our method with the state-of-the-art methods in R-Precision, Frechet Inception Distance (FID), Multimodal Distance (MM-Dist) and Multimodality (MModality) metrics.
Our Diverse-T2m outperforms the state-of-the-art methods in most key metrics, \emph{i.e.}, ranks first for metrics R-Precision~(Top-1, Top-2, Top-3), MM-Dist, MModality and ranks second for FID. 

MModality reflects the ability of generating diverse motions based on a text specified by a user, and we can find that our diverse-T2M substantially outperforms all previous methods in this metric. The previous best performing method for generating diversity was MDM (MultiModality 2.799), which we outperform by 43 percent. MDM generates a relatively high diversity of motions, but the text has less control over the motions, which is reflected in the metric MultiModal Dist, which is almost twice as worse than ours. This phenomenon is also present in R Precision. Compared to previous methods that generate higher diversity, we have a stronger ability to diversify the generation, and in addition to this, our method has stronger textual control as well as more realism for the generated motions.

For previous diffusion-based methods, such as ReMoDiffuse, MotionLCM , Motion Mamba, we outperform these models with higher motion-text matching accuracy, \textit{i.e.}, we gain significant improvement (+1.5 Top-1, + 1.9 Top-2, +1.4 Top-3) on the best ReMoDiffuse of the three. Not only do we outperform them all in the semantic consistency metrics, \textit{i.e.}, R Precision and MultiModal Dist, but we are better at generating diversity, such as outperform ReMoDiffuse 121\% and Motion Mamba by 73\% in MultiModality.
In addition, these models often exhibit slower inference speed due to the iterative nature of diffusion processes.

In contrast to MoMask, which possess best ability of text control during generation, our diverse-T2M not only do we outperform them all in the semantic consistency metrics they specialize in, but we are better at generating diversity. In details, we gain the improvement (+0.4 Top-1, + 0.4 Top-2, +0.2 Top-3) on MoMask and outperform it by 220\% in MultiModality.

\textbf{KIT-ML}.
We further evaluate our method on the commonly used large-scale dataset KIT-ML.
As the results in Table~\ref{tab:table2} shows, our Diverse-T2m outperforms the state-of-the-art methods in key metrics, \emph{i.e.}, ranks first for metrics R-Precision~(Top-1), MModality and ranks second for R-Precision~(Top-2, Top-3) and MultiModal Dist. 

The results on the KIT-ML dataset indicate that our diverse-T2M is at a high level in the performance of semantic consistency, which proves that we can generate semantically well-fitting motions with a textual description. We have the highest generation diversity, much higher than other methods, while maintaining the ability to maintain excellent text control, such as outperform ReMoDiffuse 179\%, 206\% and Motion Mamba by 106\% in MultiModality.

\begin{figure*}[t]
\centering
\includegraphics[width=0.8\textwidth]{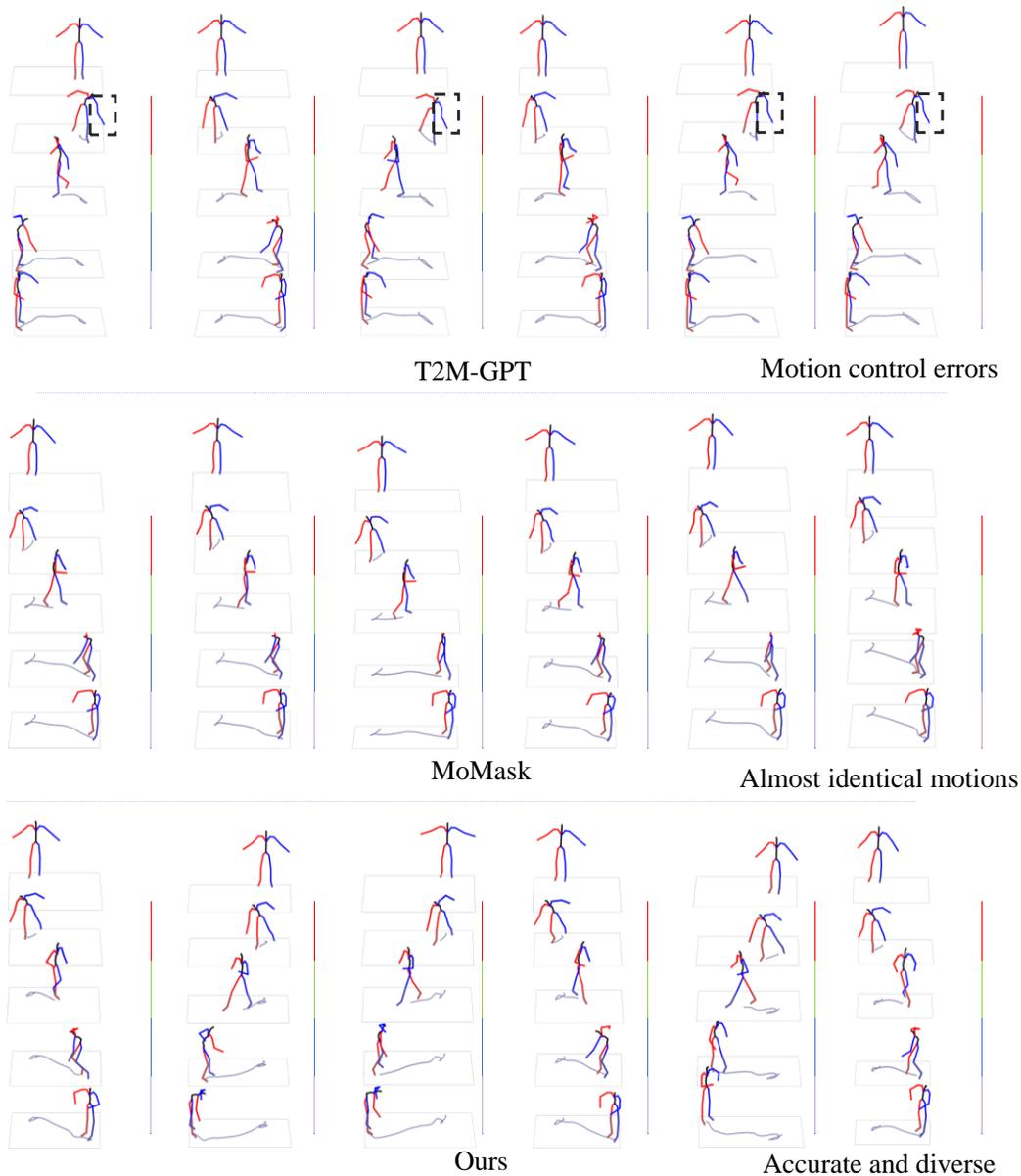}
\caption{Qualitative comparison. We compare our method qualitatively with T2M-GPT~\cite{zhang2023generating} and MoMask~\cite{guo2024momask}, both follow the two-stage paradigm. 
The other two methods have their drawbacks: T2M-GPT experiences control failures, while MoMask exhibits homogenization in its generation results.
In contrast, our Diverse-T2M not only achieves strong text-motion semantic consistency but also exhibits significant diversity. For example, it varies the direction of walking, the hand used for the second pick, and the posture when wiping.
}
\label{figure6}
\end{figure*}

\begin{figure*}[t]
\centering
\includegraphics[width=1\textwidth]{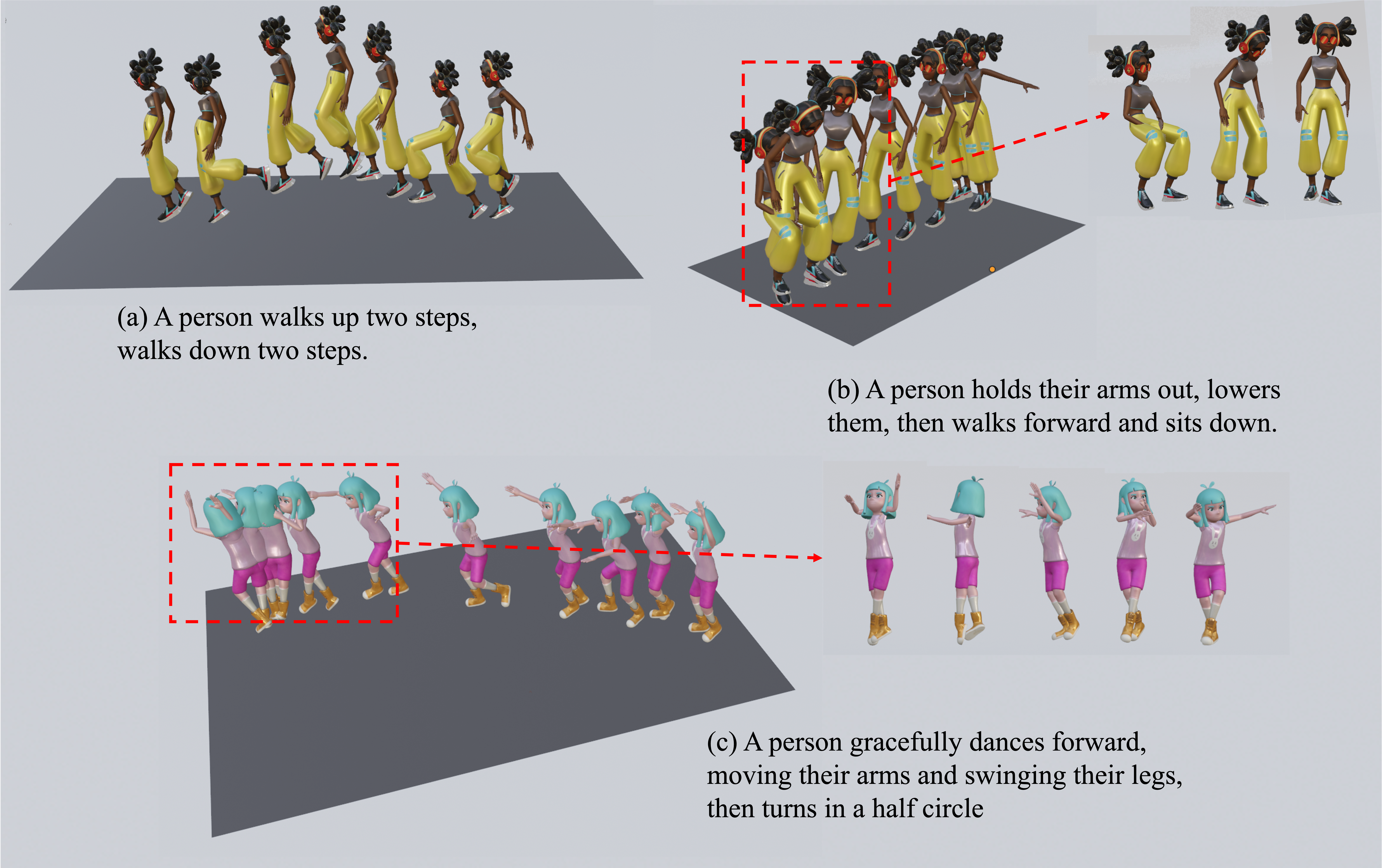}
\caption{Visual results of our Diverse-T2M can generate precise and high-fidelity human motion consistent with the given text descriptions.
}
\label{figure7}
\end{figure*}

\begin{figure*}[t]
\centering
\includegraphics[width=1\textwidth]{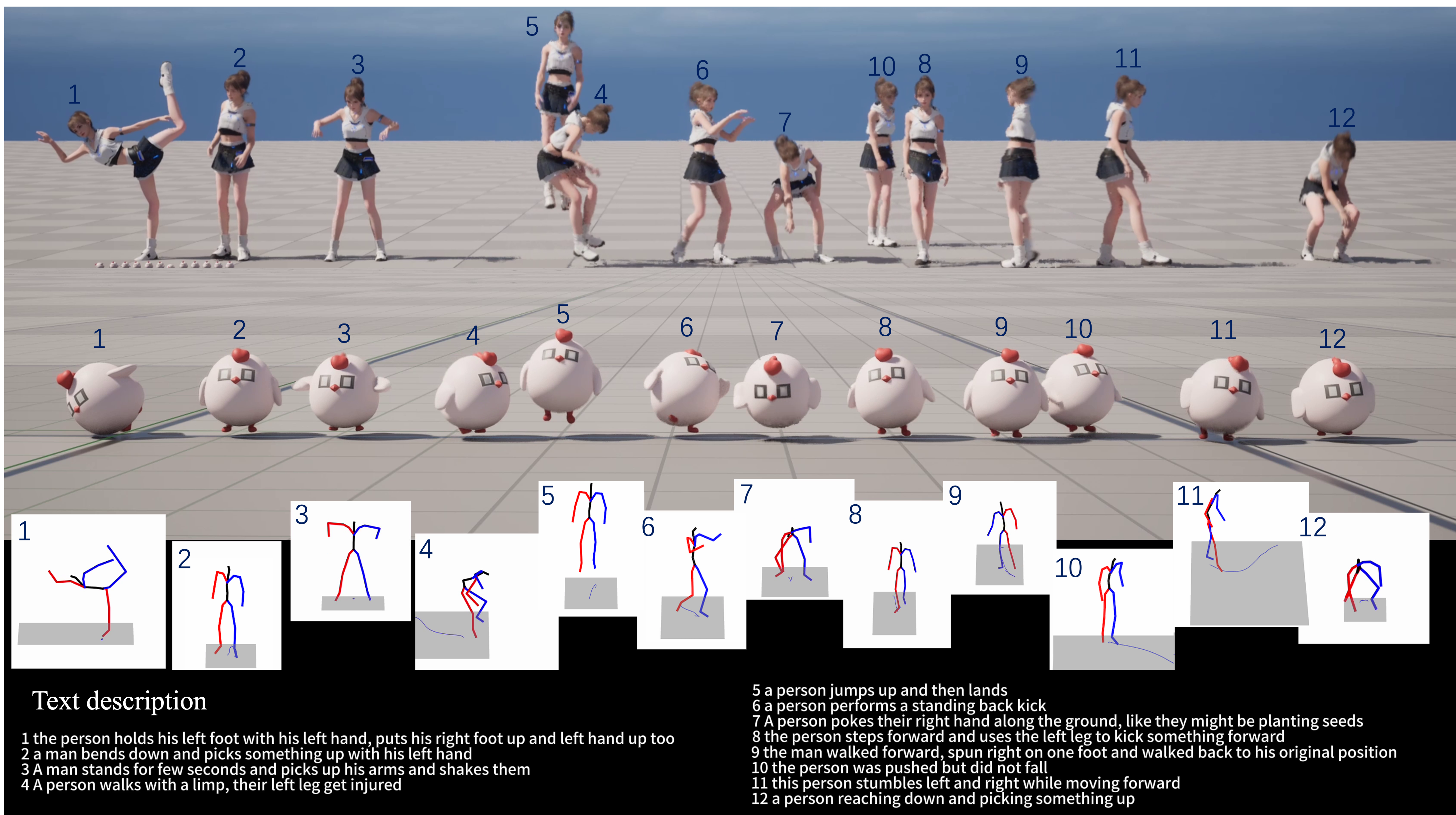}
\caption{
Illustration of Diverse-T2M's application in a business scenario. With our Diverse-T2M, animators only need to input the required text~(bottom) and multiple animations consistent with the text are automatically generated. Additionally, non-human character chick can be animated by rigging the chick character with a human skeletal structure, demonstrate the potential for adapting non-human character-animated.
}
\label{figure8}
\end{figure*}

\subsection{Ablation Study}
We conduct a series of ablative experiments to evaluate the performance contribution of each component of our proposed Diverse-T2M.

Baseline setting: 1) Motion discrete representation: Since for the discrete representation of the motion is used the technique community already has the technique VQ-VAE or RVQ-VAE, we perform ablation experiments on both baselines.
2) classifier-free guidancer strategy: following~\cite{guo2024momask, wang2023t2m}, they set $\emptyset$ to some text, we make this as our baseline for verify the effectiveness of our proposed motion generation under noise and text conditions.

\textbf{Effectiveness of Variational Text-to-codes Predictor.}
As shown in Table~\ref{table3}, our designed Variational Text-to-codes Predictor achieves performance improvements over both VQ-VAE and RVQ-VAE baseline.
In details, it brings a boost in generation diversity~(+0.774 MultiModality for VQ-VAE setting and + 0.67 MultiModality for RVQ-VAE setting). In the Variational Text-to-codes Predictor, we construct a hidden space within transformer structure, changing the original one-to-one mapping to map a data point to a distribution in the hidden space. Furthermore, we design a sampler that maps to a probability distribution of the motion code after sampling the distribution corresponding to the text in the hidden space. These designs can effectively overcome the problem of poor diversity of mapping-based generation methods despite their high control accuracy. In previous transformer-based methods, a text can only be mapped to a point in the feature space, whereas we can map a text to a data distribution in the hidden space. These are more in line with the objective laws of text-motion correspondence in the real world: due to individual differences, a sentence describing an motion should have a large solution space containing a large number of solutions that are different but consistent with the textual description.
In addition to diversity, there are improvements in other areas, such as in the metrics R-precision and MultiModal Dist reflecting semantic consistency~(+0.4 Top1, +0.2 Top2, +0.1 Top3 for VQ-VAE setting and +0.6 Top1, +0.2 Top2, +0.3 Top3 for RVQ-VAE setting), and in the metrics reflecting the quality of the reconstruction FID over the baseline. This reflects the superiority of the modules we designed.

\textbf{Effectiveness of Noise Signal.}
Table~\ref{table3} shows that the noise signal can significantly improve the generation diversity while maintaining a high level of text control accuracy(+2.143 MultiModality for VQ-VAE setting and + 2.085 MultiModality for RVQ-VAE setting).
Diversity must be carried by diverse information, and noise can assume such a random source of information, for example, in diffusion-based methods, the input noise is sampled randomly, which keeps the generated images diverse.
As an example, when we design diverse-T2M to generate thirty motions for the same text that match its description, although the text signals are all the same, the noise signals are different for each inference, and the noise carries a diversity of information carriers.
As an example, when we design diverse-T2M to generate thirty motions for the same text that match its description, although the text signals are all the same, the noise signals are different for each inference, and the noise carries the diversity of information carriers, which enables the thirty inference to generate motions that are diverse and match the text description.

In addition, we conducted experiments with fusion ratio settings for distributions generated from text and noise signals. In Figure~\ref{figure5}, The larger $w$ is, the more the noise signal affects the final generated motion. Digram (b) indicates that the generation diversity is continuously getting larger as $w$ increases.
Analyzing digram (a,c,d) reveals that as $w$ increases, it causes both retrieval accuracy and multimodal distance performance to deteriorate, due to the decreasing control of the text. 
The experiments with the $w$ parameter reflect that the role of both the text signal and the noise signal on the accuracy of motion control and the diversity of motions, and that a balance between the two can be satisfied by generating both motions that conform to the text description and a variety of motions.

\subsection{Analysis of Diversity of Motion Generation}
\textbf{Qualitative Comparison.} As shown in Figure~\ref{figure6}, the text we use for qualitative comparison is a long text, which contains both a complex action flow and rich control details, such as ``use the right hand'', \textit{etc.}, which is a great test of the control of motion generation. We tested our method against two other sota methods used for comparison, using a given text as input and letting each method generate thirty motions and visualize six randomly selected motions.
Many control lapses can be seen in the T2mM-GPT, such as using the left hand where the right hand should be used, as shown in the black dashed box in the figure. The MoMask method, on the other hand, although the motions are all exact, each of them is almost identical and has a very low diversity.
In contrast, our approach can both conform to textual descriptions where the text controls and play arbitrarily within reasonable limits where the text does not.

\textbf{Analysis.} The reason why our method produces diverse results for multiple inference of the same text is because the noise signal is different for each inference, which carries the information needed for diversity, and secondly, the text and noise signals are mapped to the data distribution in latent space for each inference, and the sampling is randomized each time. Together, these affect the generative diversity of our Diverse-T2M.

\subsection{Visualization Results}
We provide some result in Figure~\ref{figure7}. It can be seen that our approach can clearly generate 3D spatial movement of positions, such as going up and down the steps in example (a); as well as movements with interaction with the outside world, such as sitting down in (b); and precision control, with only a half-turn in (c). These examples demonstrate that we maintain a high level of textual control.

\subsection{Applications}
Here we present the application in a business scenario, as shown in Figure~\ref{figure8}. After completing character modeling for the application scenario, animators often need to manually animate the character frame-by-frame, which is resource-intensive. With our Diverse-T2M, animators only need to input the required text, as illustrated at the bottom of Figure~\ref{figure8}, and multiple animations consistent with the text can be automatically generated. These animations can then be selected or further refined to meet specific business needs, thus greatly reducing the cost of animation production.

Additionally, for non-human animation requirements, the motions generated by models trained on human text-to-motion datasets can still be effectively utilized. As illustrated in the chick character in Figure~\ref{figure8}, by rigging the chick character with a human skeletal structure, anthropomorphic motions can be driven using human skeletal sequences.

When the text-to-motion model exhibits greater diversity, it can enhance the visual richness and dynamism of application scenes, preventing homogenization. Furthermore, it offers a broader array of foundational options for animators to perform secondary processing. The demo videos are provided in the supplemental materials.

\label{qua_results}

\section{Conclusion} \label{sec:conc}

We propose Diverse-T2M for text-to-motion generation, which introduces uncertainty into the generation process to strength the ability of human motion generation diversity.
Our results on text-to-motion generation benchmark datasets demonstrate that our method significantly enhances diversity while maintaining state-of-the-art performance in text consistency. By proposing the use of noise signals as carriers of diversity and constructing a latent space that encodes text into a probabilistic representation, we effectively introduce uncertainty into the generation process.
Additionally, we demonstrate the usefulness of our approach in real-world business, as well as the potential for adapting non-human character animation.



\ifCLASSOPTIONcaptionsoff
  \newpage
\fi

\bibliographystyle{IEEEtran}
\bibliography{IEEEabrv,referemce}

\begin{IEEEbiography}
[{\includegraphics[width=1in,height=1.25in,clip,keepaspectratio]{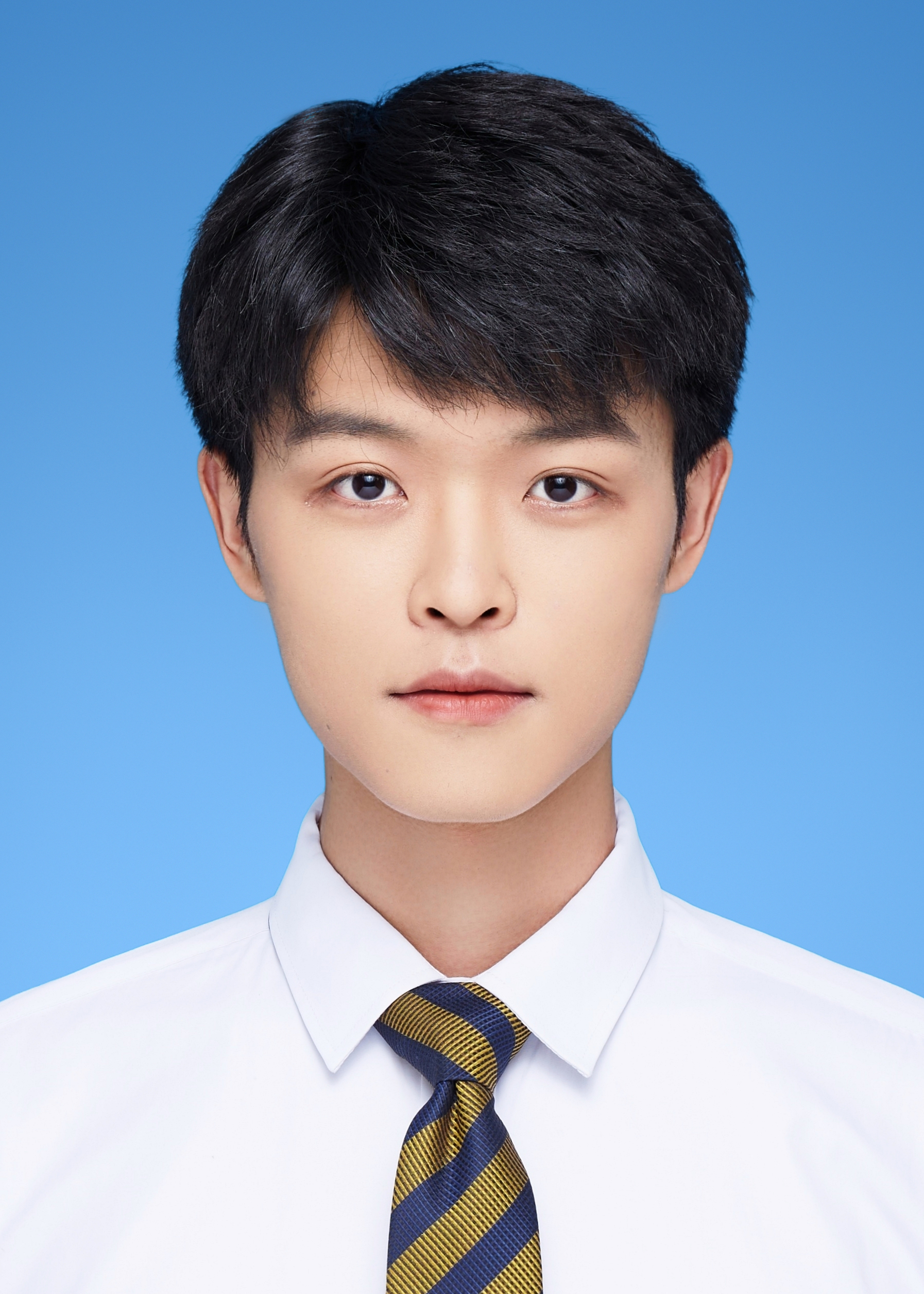}}]{Zheng Qin}
received the B.S. degree in robotic engineering from the Harbin Institute of Technology, China, in 2021. He is currently working toward the PdD. degree in artificial intelligence from Xi'an Jiaotong University. His research interests include multi-object tracking, visual navigation, motion generation and video generation.
\end{IEEEbiography}

\begin{IEEEbiography}
[{\includegraphics[width=1in,height=1.25in,clip,keepaspectratio]{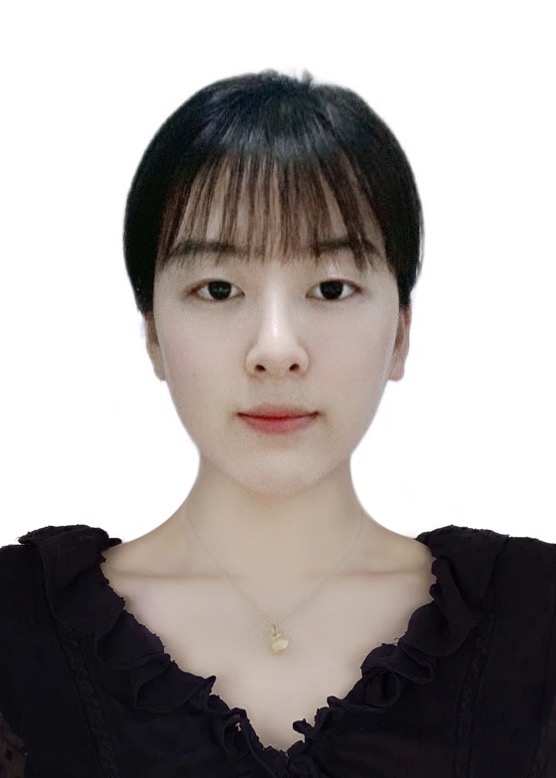}}]{Yabing Wang}
received the BE degree in software engineering from Zhengzhou University, Zhengzhou, China, in 2020, and the ME degree in College of Computer Science and Technology from Zhejiang Gongshang University, Hangzhou, China, in 2023. She is currently working toward the PhD degree with the Institute of Artificial Intelligence and Robotics, Xi’an Jiaotong University. Her research interests include multi-modal learning.
\end{IEEEbiography}

\begin{IEEEbiography}[{\includegraphics[width=1in,height=1.25in,clip,keepaspectratio]{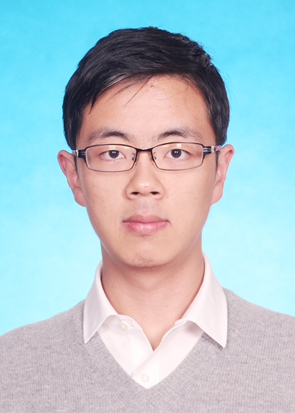}}]{Minghui Yang} received the B.S. degree in Computer Science from Shanghai Jiao Tong University in 2015. He is currently an Algorithm Engineer at Ant Group, responsible for digital human algorithms. His research interests include 3D/2D digital humans, dialogue, and multimodal interaction.
\end{IEEEbiography}

\begin{IEEEbiography}[{\includegraphics[width=1in,height=1.25in,clip,keepaspectratio]{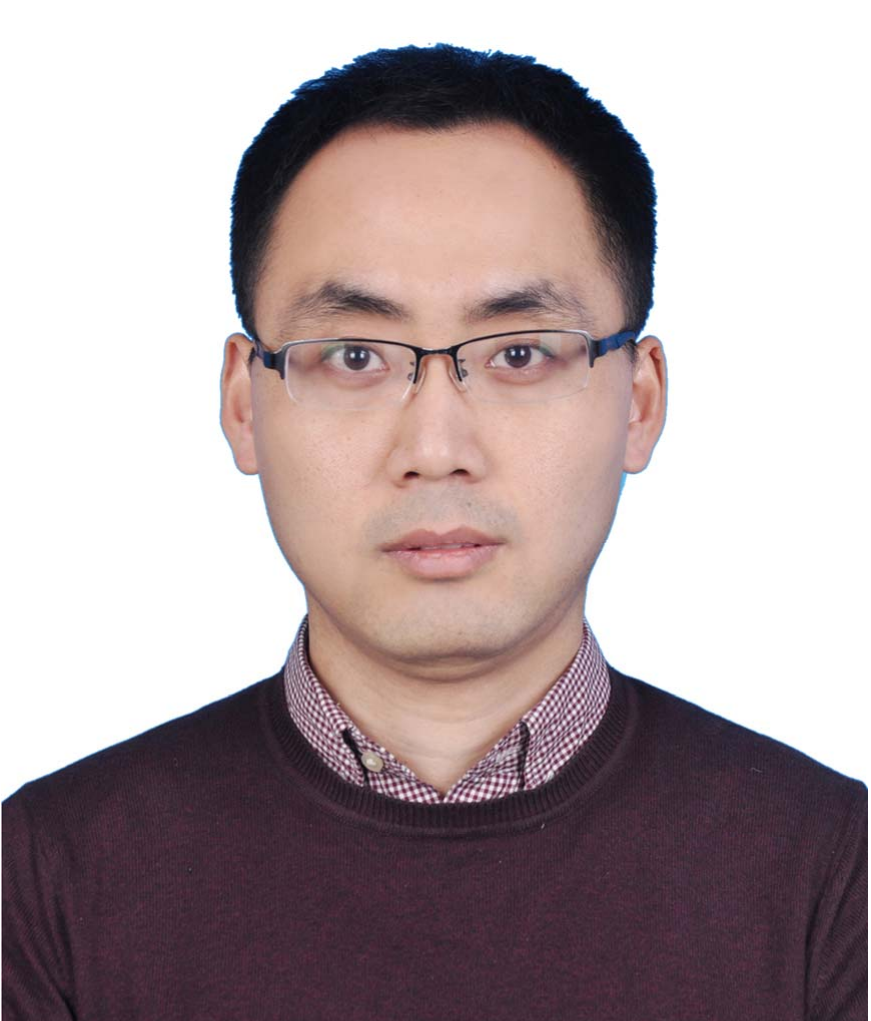}}]{Sanping Zhou} (Member, IEEE)
	received Ph.D. degree in control science and engineering from Xian Jiaotong University, Xi'an, China, in 2020. From 2018 to 2019, he was a visiting Ph.D. student at the Robotics Institute, Carnegie Mellon University. He is currently an Associate Professor with the Institute of Artificial Intelligence and Robotics, Xian Jiaotong University, Xi'an, China. His research interests include machine learning and computer vision, with a focus on object detection, image segmentation, visual tracking, multitask learning, and metalearning.
\end{IEEEbiography}

\begin{IEEEbiography}[{\includegraphics[width=1in,height=1.25in,clip,keepaspectratio]{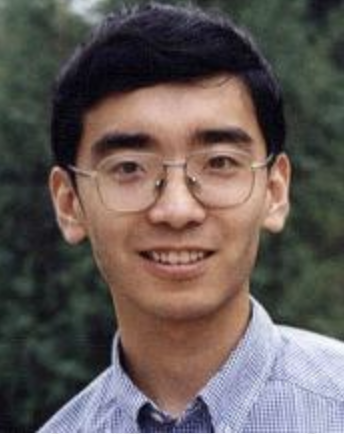}}]{Ming Yang} (Member, IEEE) received the BE and
ME degrees in electronic engineering from Tsinghua University, Beijing, China, in 2001 and 2004, respectively, and the PhD degree in electrical
and computer engineering from Northwestern University, Evanston, Illinois, in 2008. From 2004 to 2008, he was a research assistant with the Computer Vision Group, Northwestern University. After his graduation, he joined NEC Laboratories America, Cupertino, California, where he was a senior researcher. He was a research scientist in AI research at Facebook from 2013 to 2015. He is currently the
co-founder and the VP of software with Horizon Robotics, Inc. He has
authored more than 60 peer-reviewed publications in prestigious international journals and conferences, which have been cited more than
13,100 times. His research interests include computer vision, machine
learning, face recognition, large scale image retrieval, and intelligent
multimedia content analysis
\end{IEEEbiography}

\begin{IEEEbiography}[{\includegraphics[width=1in,height=1.25in,clip,keepaspectratio]{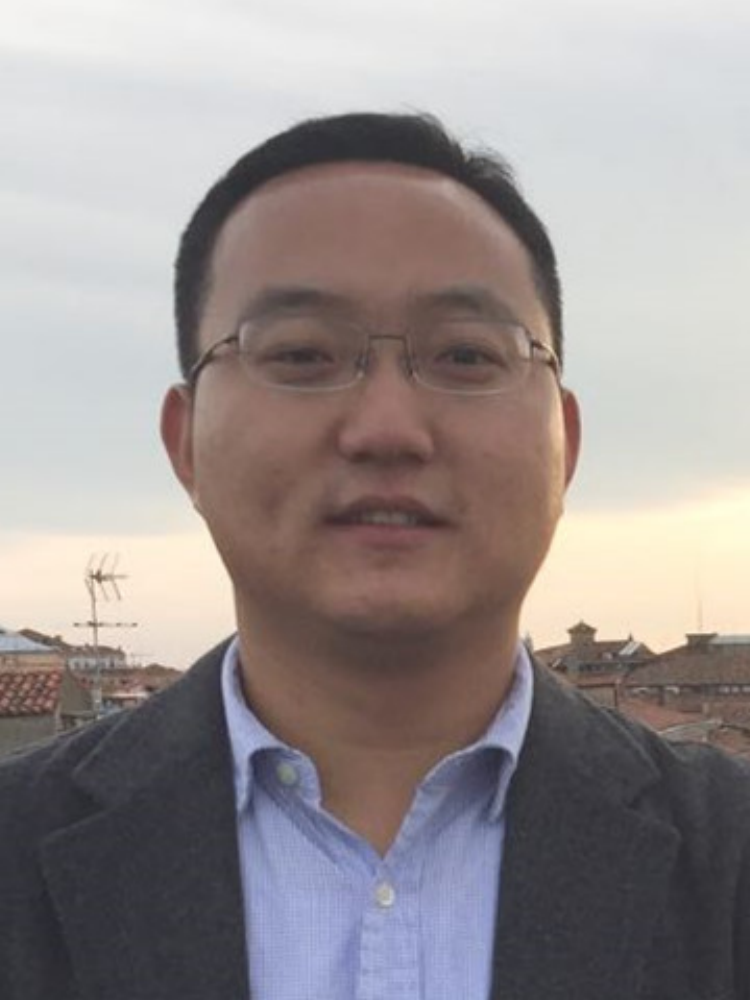}}]{Le Wang}(Senior Member, IEEE) received the B.S. and Ph.D. degrees in Control Science and Engineering from Xi'an Jiaotong University, Xi'an, China, in 2008 and 2014, respectively. From 2013 to 2014, he was a visiting Ph.D. student with Stevens Institute of Technology, Hoboken, New Jersey, USA. From 2016 to 2017, he is a visiting scholar with Northwestern University, Evanston, Illinois, USA. He is currently a Professor with the Institute of Artificial Intelligence and Robotics of Xi'an Jiaotong University, Xi'an, China. His research interests include computer vision, pattern recognition, and machine learning.
\end{IEEEbiography}




\end{document}